# ESGBERT: LANGUAGE MODEL TO HELP WITH CLASSIFICATION TASKS RELATED TO COMPANIES' ENVIRONMENTAL, SOCIAL, AND GOVERNANCE PRACTICES

Srishti Mehra*, Robert Louka*, Yixun Zhang

University of California Berkeley, School of Information, USA

## ABSTRACT

*Environmental, Social, and Governance (ESG) are non-financial factors that are garnering attention from investors as they increasingly look to apply these as part of their analysis to identify material risks and growth opportunities. Some of this attention is also driven by clients who, now more aware than ever, are demanding for their money to be managed and invested responsibly. As the interest in ESG grows, so does the need for investors to have access to consumable ESG information. Since most of it is in text form in reports, disclosures, press releases, and 10-Q filings, we see a need for sophisticated natural language processing (NLP) techniques for classification tasks for ESG text. We hypothesize that an ESG domain specific pre-trained model will help with such and study building of the same in this paper. We explored doing this by fine-tuning BERT's pre-trained weights using ESG specific text and then further fine-tuning the model for a classification task. We were able to achieve accuracy better than the original BERT and baseline models in environment-specific classification tasks.*

## KEYWORDS

*ESG, NLP, BERT, Universal Sentence Encoder, Deep Averaging Network*

## 1. INTRODUCTION

The importance of Environmental, Social, and Governance (ESG) issues has risen in prominence over the last decade. In the early 1990's, fewer than 20 publicly listed companies issued reports that included ESG data; that number grew to almost six thousand by 2014 [1]. Regulations for SEC filings to follow certain standards for responsibility about Climate Change and Human Governance, and Investor and Shareholder support has driven the motivation for these disclosures.

There has been little research analyzing the non-financial information content [2] in financial disclosures. The most common methods for analyzing the non-financial, narrative information content remain manual or dictionary-based [3][4][5][6][7][8]. The existing literature focuses on the quantity of non-financial information published, rather than its content [9][10]. This underpins the requirement for a study like ours.

Domain-specific BERT variations like FinBERT [11] and BioBERT [12], that have been either fine-tuned or pre-trained on domain corpus instead of or in addition to the generic English language, have achieved great results in studying the information content, particularly for domain specific language tasks. The primary interest of this research is to harness that benefit for ESG specific text classification tasks. We study building an environment-specific variation of BERT by fine-tuning the pre-trained BERT weights using a Masked Language Model (MLM) task on an ESG corpus and then further fine-tuning our model for Sequence Classification to predict:

1. A change or no change in environmental scores, and
2. A positive or negative change (if any) in environmental scores of companies using ESG related text in their 10-Q filings.

## 2. BACKGROUND

Accounting for Sustainability (https://www.accountingforsustainability.org/) is a project that aims to inspire action by finance leaders to drive a fundamental shift towards resilient business models and a sustainable economy. To do so the project publishes guides, case studies, blogs, reports and surveys, and hosts webinars. This material is available on their knowledge hub and is reflective of the opportunities and risks posed by environmental and social issues. These are what we chose as our ESG corpus to pre-train our BERT model on top of the English Wikipedia and BooksCorpus it has been trained on [13].

In 2010, the SEC published an interpretive release on climate change-related disclosures "to remind companies of their obligations under existing federal securities laws and regulations to consider climate change and its consequences as they prepare disclosure documents" [14]. Therefore, the company's disclosures should not only consist of financial narratives but also contain information about the environmental aspects concerning the firm [9]. This is the reason we chose to use 10-Q filings as our input for the classification task.

Sustainalytics' ESG Risk Ratings measure a company's exposure to industry-specific material ESG risks and how well a company is managing those risks. This multi-dimensional way of measuring ESG risk combines the concepts of management and exposure to arrive at an absolute assessment of ESG risk. These risk scores are also broken down into environmental, social, and governance risks. Of these, for our research, we use the change in total environmental risk score for each company quarter over quarter to indicate whether there was: 1. A change or no change, and 2. A positive or negative change.

## 3. RELATED WORK

This section describes previous research conducted on domain-specific variations of BERT (3.1) and ESG related NLP research (3.2).

### 3.1. Domain-specific BERT variants

FinBERT [11] author explored pre-training BERT on Financial corpus based on their learning from a previous study by Howard and Ruder [15] which shows that further pre-training a language model on a target domain corpus improves the eventual classification performance. They pretrained BERT on finance-specific corpora and used those weights to further train the model for financial sentiment classification. They saw improved results in comparison to the original BERT (pre-trained on generic English language corpora).

BioBERT [12] authors, similarly (similar architecture as followed by FinBERT), pre-trained BERT with Biomedical corpora in addition to the English language corpora it was already trained on. They went on to find that BioBERT largely outperformed BERT in a variety of biomedical text mining tasks.

### 3.2. ESG related NLP research

Armbhurst, Schäfer, and Klinger, 2020 studied the effect of the environmental performance of a company (as learned from MD&A sections in 10-K and 10-Q filings) on the relationship between the company's disclosures and financial performance. They found that textual information contained within the MD&A section does not allow for conclusions about the future (corporate) financial performance. However, there is evidence that the environmental performance can be extracted by NLP methods.

Serafeim and Yoon [1] showed that ESG ratings predict future ESG news and market reactions, particularly when there is disagreement amongst raters. This study is similar to our study in that it uses ESG scores (from TruValue, a company similar to Sustainalytics, which we use for ESG scores) to predict the public reaction, whereas we are using information from public documents to predict ESG scores. The news in their study was aggregated by TruValue using machine learning enabled text mining from a wide variety of sources.

## 4. METHOD

This section will be divided into BERT (4.1), pre-trained BERT weights on ESG specific corpus (4.2), and fine-tuning for the classification task mentioned in earlier sections (4.3).

### 4.1. Bidirectional Encoder Representations from Transformers (BERT)

BERT [13] is a pre-trained model that builds word representations learned through bi-directional tasks. They use a Masked Language Model (MLM) task to fuse the left and the right context, which allows them to pretrain a deep bidirectional Transformer. The task randomly masks some of the tokens from the input and predicts the masked words based only on context. Additionally, they use the Next Sentence Prediction (NSP) task that captures the relationship between two sentences which is not directly captured by language modeling. For finetuning, the BERT model is first initialized with the pre-trained parameters learned in the bi-directional approach, and those parameters are then fine-tuned using labeled data from the downstream tasks. Each downstream task has separate fine-tuned models, even though they are initialized with the same pre-trained parameters [13].

### 4.2. Pre-training on ESG Specific Corpus

BERT's pre-training procedure largely follows the existing literature on language model pretraining, they use the BooksCorpus (800M words) [16] and English Wikipedia (2,500M words) for the same.

Since the text in our research is also in the English language, we did not want to forgo the benefit of pre-trained weights on such large English language corpora. Thus, we further train BERT's pre-trained weights using an additional Masked Language Modeling (MLM) Task.

We use text from the Knowledge Hub of Accounting for Sustainability for our MLM task. These occur in the form of guides, case studies, blogs, reports, and surveys. We tokenized the text found in those documents using BERT's WordPiece Tokenizer, masked 15% of the words, and learned to predict those masked words. In doing so, we updated the pre-trained weights of BERT to reflect learnings from ESG context. We chose the MLM task since it learns to predict the masked words based only on context.

### 4.3. Fine-Tuning for Classification Task

In order to test our hypothesis of a domain-specific variation of BERT working better than that original trained on generic language, we chose two classification tasks that we fine-tuned on. The classification tasks were to predict whether there was:

1.      A change or no change, and
2.      A positive or negative change (if any) in environmental scores of companies using ESG related text in their 10-Q filings.

#### 4.3.1. Input for Fine-Tuning with Classification

BERT takes up to 512 tokens as input. Since our input per company per quarter was an entire 10Q document, which are multiple pages long, we needed a way to choose 512 tokens from each document. 10-Q reports contain small portions that address environmental factors. Therefore, our approach was to extract the sentences most relevant to environmental factors and choose 512 tokens from those. In order to do so, we needed a method to order all sentences in the report (or pick the top 3) by relevance. We encoded sentences in the reports and compared them using cosine similarity with a few benchmark sentences that we thought would help us extract the most relevant sentences from these documents.

For the encoding of the sentences, we experimented with Sentence BERT [17] and Universal Sentence Encoder [18]. We found that the Deep Averaging Network (DAN) version of the Universal Sentence Encoder works in this case, to extract the most relevant sentences. Since we were using the DAN version of the Universal Sentence Encoder, we created our benchmark sentence as one that was a scramble of words that reflect high relevance with environment factors. We hypothesize that the scramble of words helped us extract a deeper, more diverse set of relevant sentences from the documents.

After encoding and comparing each sentence in the report with the benchmark sentence(s), we chose the top 3 sentences for each document and fed that as input to our model. We let there be a truncation for those that exceeded 512 tokens and padding for those that had less than 512 tokens in the 3 sentences chosen.

#### 4.3.2. Fine-Tuning

The approach for fine-tuning for both the classification tasks was to use BERT embeddings and attention masks of the chosen 512 tokens and feed them into the model that was fine-tuned on the ESG corpus. The outputs of that were used as inputs for a classification layer to learn ESG scores. The outputs for the classification layer were scores from Sustainalytics for each company for each. The models were hyperparameter tuned to achieve the results discussed in section 6.

## 5. DATA

### 5.1. ESG Corpus for Fine-Tuning

The ESG corpus that we fine-tuned BERT's pre-trained weights on was obtained from the Knowledge Hub of the Accounting for Sustainability project.

## 5.2. Input and Output for Classification

For our input, we got 10-Q reports for S&P 500 companies from University of Notre Dame's Software Repository for Accounting and Finance for the time frame 2014-2018. For the output, we used Wharton's research platform WRDS to obtain quarterly Sustainalytics scores for the same companies for the same time frame.

## 5.3. EDA

The distribution of scores (Figure 1) does not vary highly. Roughly 60% of the quarterly changes in environmental scores are zero. While the tails of the distribution do contain score changes on the larger side, most of the changes are quite small. This does not affect our architecture much since we are doing binary classifications (change or no change; positive change or negative change). Figure 2 shows the relative frequency of the sentence lengths which was used to decide our truncation and padding strategy to ensure we feed our model 512 tokens each time.

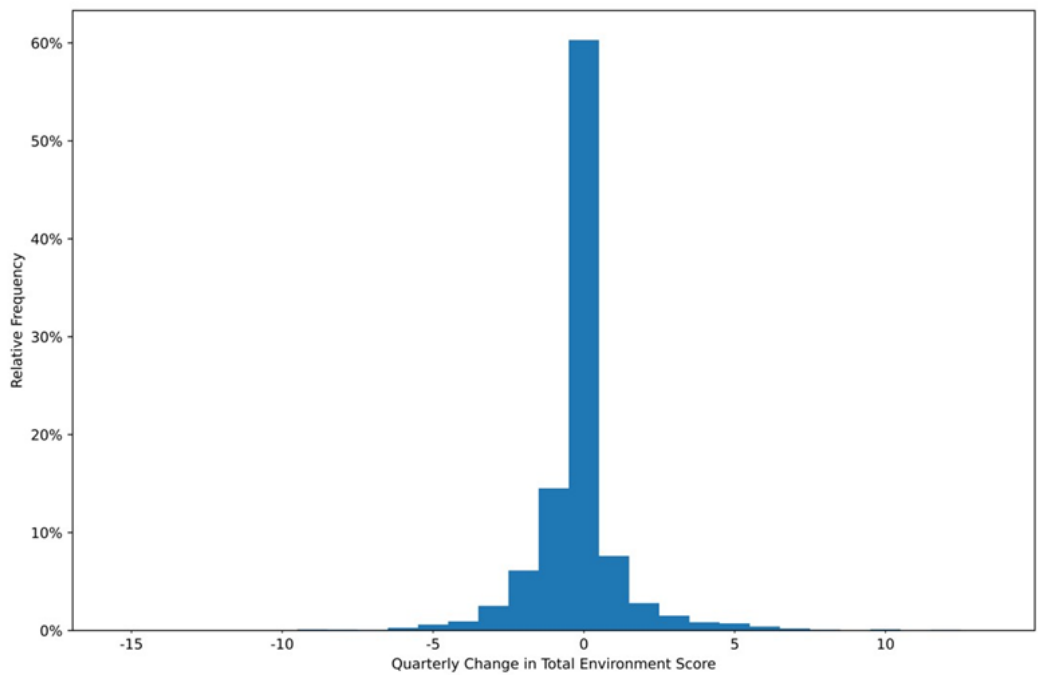

Figure 1. Quarterly change in Total Environment Scores for companies

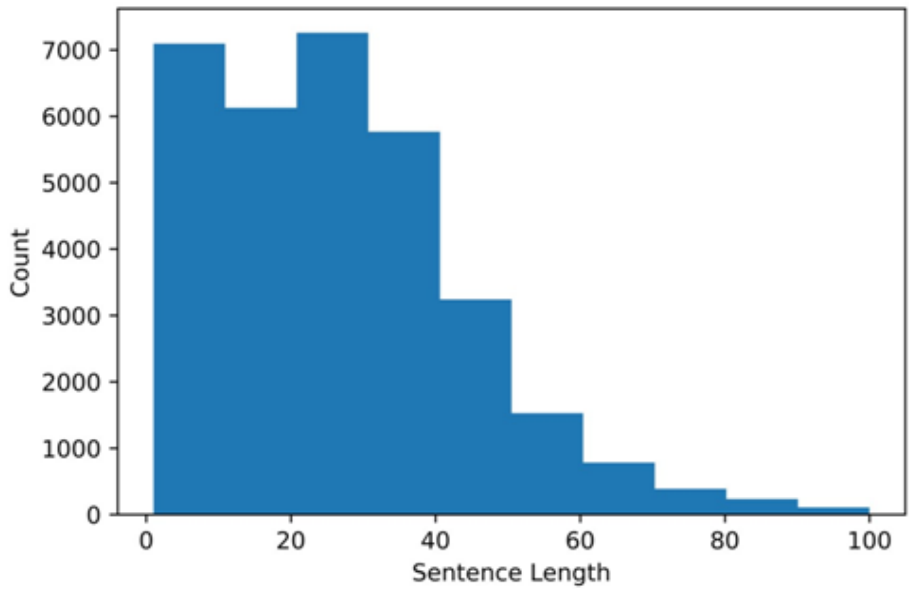

Figure 2. Histogram of length of sentences in the 10-Q filings

## 6. RESULTS

Naive Bayes barely beat the common class in training and could not beat it in the validation or test data sets. ESGBERT is able to outperform BERT and the other classification techniques we compared against in both tasks that we trained our model on. The results are illustrated in Table 1 and Table 2 respectively, and details about the relevant hyperparameters are mentioned in the description below each table.

Table 1. Classification for change or no change in environmental risk score of company per quarter. Models run with learning rate 2e-05, epsilon 1e-08, 8 epochs, and batch size of 8.

| Model | Train Accuracy | Validation Accuracy | Test Accuracy |
|---|---|---|---|
| Common Class Prediction | 0.6107 | 0.614 | 0.5791 |
| BERT | 0.6251 | 0.6325 | 0.5985 |
| ESGBERT | 0.839 | 0.7906 | 0.6709 |

Table 2. Classification for positive or negative change in environmental risk score of company per quarter. Models run with learning rate 2e-05, epsilon 1e-08, 8 epochs, and batch size of 8.

| Model | Train Accuracy | Validation Accuracy | Test Accuracy |
|---|---|---|---|
| Common Class Prediction | 0.5974 | 0.5978 | 0.5682 |
| BERT | 0.6583 | 0.6055 | 0.4317 |
| ESGBERT | 0.8618 | 0.8 | 0.793 |

## 7. CONCLUSION

With our research, we strengthen confidence in the learning that pre-training BERT on domain specific corpus yields better results in classification tasks related to that domain. We anticipate that ESGBERT's pre-trained weights, that have learned ESG context, can be used for multiple ESG specific text classification tasks and researchers/developers will benefit from them. For example, our model's pre-trained weights can be used to predict Social and Governance risk scores for companies in addition to the Environmental risk scores that we predicted.

Additionally, the weights can be enhanced by training on additional ESG corpora, like ESG disclosures that companies have now started to release. Since such disclosures will fully focus on the Environmental, Social, and Governance practices and investments, they will have more to inform about the scores than 10-Q filings did.

## AUTHOR

**Srishti Mehra, Robert Louka, and Yixun Zhang** are graduate students of University of California, Berkeley, studying/studied Masters in Information and Data Science.

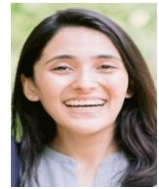